\documentclass[letterpaper, 10 pt, conference]{ieeeconf}  

\usepackage[backend=biber]{biblatex}
\usepackage{booktabs}
\usepackage{graphicx}

\usepackage{xparse}
\usepackage{amsmath}
\usepackage{amssymb}
\usepackage{color}
\usepackage{pgfplots} 
\usepackage{todonotes}

\usepackage[font=small]{caption}

\usepackage{notation_memmesheimer}
\usepackage{hyperref}
\usepackage{lipsum}
\def\andothers{et al.\ }
\def\figname{Fig.\,}

\def\secname{Section\,}
\def\tabname{Table\,}

\newcommand{\citeauthordirect}[1]{\citeauthor{#1} \cite{#1}}

\newcommand{\approachname}{Skeleton-DML}
\newcommand{\skeletondmlimpro}{+3.3\%}
\newcommand{\skeletondmlaugmentedimprosldml}{+7.7\%} 
\newcommand{\skeletondmlaugmentedimpro}{+4.4\%} 

\definecolor{drink_water}{rgb}{0.223529, 0.231373, 0.474510}
\definecolor{throw}{rgb}{0.321569, 0.329412, 0.639216}
\definecolor{tear_up_paper}{rgb}{0.419608, 0.431373, 0.811765}
\definecolor{take_off_glasses}{rgb}{0.611765, 0.619608, 0.870588}
\definecolor{reach_into_pocket}{rgb}{0.388235, 0.474510, 0.223529}
\definecolor{pointing_to_something_with_finger}{rgb}{0.549020, 0.635294, 0.321569}
\definecolor{wipe_face}{rgb}{0.709804, 0.811765, 0.419608}
\definecolor{falling}{rgb}{0.807843, 0.858824, 0.611765}
\definecolor{feeling_warm}{rgb}{0.549020, 0.427451, 0.192157}
\definecolor{hugging_other_person}{rgb}{0.741176, 0.619608, 0.223529}
\definecolor{put_on_headphone}{rgb}{0.905882, 0.729412, 0.321569}
\definecolor{hush_(quite)}{rgb}{0.905882, 0.796078, 0.580392}
\definecolor{staple_book}{rgb}{0.517647, 0.235294, 0.223529}
\definecolor{sniff_(smell)}{rgb}{0.678431, 0.286275, 0.290196}
\definecolor{apply_cream_on_face}{rgb}{0.839216, 0.380392, 0.419608}
\definecolor{open_a_box}{rgb}{0.905882, 0.588235, 0.611765}
\definecolor{arm_circles}{rgb}{0.482353, 0.254902, 0.450980}
\definecolor{yawn}{rgb}{0.647059, 0.317647, 0.580392}
\definecolor{grab_other_persons_stuff}{rgb}{0.807843, 0.427451, 0.741176}
\definecolor{take_a_photo_of_other_person}{rgb}{0.870588, 0.619608, 0.839216}

\ExplSyntaxOn
\NewDocumentCommand{\storereviewer}{mm}
  {
   \bcp_store_data:nn { #1 } { #2 }
  }

\DeclareExpandableDocumentCommand{\getreviewer}{O{1}m}
 {
  \bcp_get_data:nn { #1 } { #2 }
 }

\cs_new_protected:Npn \bcp_store_data:nn #1 #2
 {
  \seq_if_exist:cF { l_bcp_data_#1_seq } { \seq_new:c { l_bcp_data_#1_seq } }
  \seq_set_split:cnn { l_bcp_data_#1_seq } { } { #2 }
 }
\cs_generate_variant:Nn \seq_set_split:Nnn { c }

\cs_new:Npn \bcp_get_data:nn #1 #2
 {
  \seq_item:cn { l_bcp_data_#2_seq } { #1 }
 }
\ExplSyntaxOff

\usepackage{pgfkeys}
\pgfkeys{
 /points array/.is family, /points array,
 .unknown/.style = {\pgfkeyscurrentname/.initial = #1},
}

\newcommand\resultsstorage[1]{\pgfkeys{/points array, #1}}
\newcommand\results[1]{\pgfkeysvalueof{/points array/#1}}

\resultsstorage{apsr20 = 29.1, apsr40 = 34.8, apsr60 = 39.2, apsr80 = 42.8, apsr100 = 45.3,
                caetanoorientation20 = 00.0, caetanoorientation40 = 00.0, caetanoorientation60 = 00.0, caetanoorientation80 = 00.0, caetanoorientation100 = 34.4,
                caetanomagnitude20 = 00.0, caetanomagnitude40 = 00.0, caetanomagnitude60 = 00.0, caetanomagnitude80 = 00.0, caetanomagnitude100 = 44.4,
                caetanomagnitudeorientation100=39.2,
                tssi100 = 41.0,
                skepxel100 = 29.6,
                gimme_signals100 = 41.5,
                sabater100 = 46.5,
                sldml20 = 36.7, sldml40 = 42.4, sldml60 = 49.0, sldml80 = 46.4, sldml100 = 50.9,
                sldmlreindex20 = 28.6, sldmlreindex40 = 37.5, sldmlreindex60 = 48.6, sldmlreindex80 = 48.0, 
                sldmlreindex100 = 54.2,
                sldmlswap_transform_default_emb_128_loss_ms_miner_ms_alpha_1.0_beta_0.0 = 55.2,
                sldmlswap_transform_default_emb_256_loss_ms_miner_ms_alpha_1.0_beta_0.0 = 50.6,
                sldmlswap_transform_default_emb_512_loss_ms_miner_ms_alpha_1.0_beta_0.0 = 52.7,
                sldmlswap_transform_default_emb_128_loss_ms_miner_triplet_margin_alpha_1.0_beta_0.0 = 51.5,
                sldmlswap_transform_default_emb_256_loss_ms_miner_triplet_margin_alpha_1.0_beta_0.0 = 51.7,
                sldmlswap_transform_default_emb_512_loss_ms_miner_triplet_margin_alpha_1.0_beta_0.0 = 54.0,
                sldmlswap_transform_rotation_5_emb_128_loss_ms_miner_ms_alpha_1.0_beta_0.0 = 51.8,
                sldmlswap_transform_rotation_5_emb_256_loss_ms_miner_ms_alpha_1.0_beta_0.0 = 55.3,
                sldmlswap_transform_rotation_5_emb_512_loss_ms_miner_ms_alpha_1.0_beta_0.0 = 55.8,
                sldmlswap_transform_rotation_5_emb_128_loss_ms_miner_triplet_margin_alpha_1.0_beta_0.0 = 53.6,
                sldmlswap_transform_rotation_5_emb_256_loss_ms_miner_triplet_margin_alpha_1.0_beta_0.0 = 54.8,
                sldmlswap_transform_rotation_5_emb_512_loss_ms_miner_triplet_margin_alpha_1.0_beta_0.0 = 55.5,
                sldmlreindex_transform_default_emb_128_loss_ms_miner_ms_alpha_1.0_beta_0.0 = 54.7,
                sldmlreindex_transform_default_emb_256_loss_ms_miner_ms_alpha_1.0_beta_0.0 = 51.5,
                sldmlreindex_transform_default_emb_512_loss_ms_miner_ms_alpha_1.0_beta_0.0 = 53.1,
                sldmlreindex_transform_default_emb_128_loss_ms_miner_triplet_margin_alpha_1.0_beta_0.0 = 47.5,
                sldmlreindex_transform_default_emb_256_loss_ms_miner_triplet_margin_alpha_1.0_beta_0.0 = 51.9,
                sldmlreindex_transform_default_emb_512_loss_ms_miner_triplet_margin_alpha_1.0_beta_0.0 = 54.0,
                sldmlreindex_transform_rotation_5_emb_128_loss_ms_miner_ms_alpha_1.0_beta_0.0 = 55.3,
                sldmlreindex_transform_rotation_5_emb_256_loss_ms_miner_ms_alpha_1.0_beta_0.0 = 58.0,
                sldmlreindex_transform_rotation_5_emb_512_loss_ms_miner_ms_alpha_1.0_beta_0.0 = 58.6,
                sldmlreindex_transform_rotation_5_emb_128_loss_ms_miner_triplet_margin_alpha_1.0_beta_0.0 = 56.0,
                sldmlreindex_transform_rotation_5_emb_256_loss_ms_miner_triplet_margin_alpha_1.0_beta_0.0 = 55.1,
                sldmlreindex_transform_rotation_5_emb_512_loss_ms_miner_triplet_margin_alpha_1.0_beta_0.0 = 56.1,
                }
\usepackage[detect-none]{siunitx}
\pgfplotsset{width=.8\linewidth,compat=1.9}

\begin{document}

\title{Skeleton-DML: Deep Metric Learning for Skeleton-Based One-Shot Action Recognition}

\IEEEoverridecommandlockouts
\author{Raphael Memmesheimer \and  Simon H\"aring \and Nick Theisen \and Dietrich Paulus
    \thanks{All authors are with the Active Vision Group, Institute for Computational Visualistics, University of Koblenz-Landau, Germany}%
    \thanks{Corresponding email: raphael@uni-koblenz.de}
}

\maketitle

\begin{abstract}
One-shot action recognition allows the recognition of human-performed actions with only a single training example. This can influence human-robot-interaction positively by enabling the robot to react to previously unseen behaviour.
We formulate the one-shot action recognition problem as a deep metric learning problem and propose a novel image-based skeleton representation that performs well in a metric learning setting. Therefore, we train a model that projects the image representations into an embedding space. In embedding space similar actions have a low euclidean distance while dissimilar actions have a higher distance. The one-shot action recognition problem becomes a nearest-neighbor search in a set of activity reference samples.
We evaluate the performance of our proposed representation against a variety of other skeleton-based image representations. In addition we present an ablation study that shows the influence of different embedding vector sizes, losses and augmentation. Our approach lifts the state-of-the-art by \skeletondmlimpro{} for the one-shot action recognition protocol on the NTU RGB+D 120 dataset under a comparable training setup. With additional augmentation our result improved over \skeletondmlaugmentedimprosldml.
\end{abstract}

\IEEEpeerreviewmaketitle

\section{Introduction}
Action recognition is a research topic that is applicable in many fields like surveillance, human robot interaction or in health care scenarios. In the past, a strong research focus was laid on the recognition of known activities, whereas learning to recognize from few samples gained popularity only recently \cite{liu2019ntu, memmesheimer2020signal}.

Because of RGB-D cameras availability and wide mobile indoor applicability, indoor robot systems are often equipped with them \cite{pages2016tiago, yamamoto2018human}.
RGB-D cameras that support the OpenNI SDK not only provide color and depth streams but also provide human pose estimates in form of skeleton sequences. These skeleton estimates allow a wide variety of higher-level applications without investing in the human pose estimation problem. As the pose estimation approach is based on depth streams \cite{zhang2012microsoft}, it is robust against background information as well as different lighting conditions and therefore also remains functional in dark environments. 
Especially in a robotics context one-shot action recognition enables a huge variety of applications to improve the human-robot-interaction. A robot could initiate a dialog, when recognizing an activity that it is unfamiliar with, in order to assign a robot-behavior to the observation. This can be done with a single reference sample, while standard action recognition approaches can only recognize actions that were given during training time. In our proposed one-shot action recognition approach observations are projected to an embedding space in which similar actions have a low distance and dissimilar actions have a high distance. 
A high distance to all known activities can be seen as an indicator for anomalies.
The embedding in a metric learning setting allows online association of novel observations which is a high advantage over classification tasks that would require retraining or fine-tuning.

\begin{figure}
    \vspace{0.06in}
    \centering
    \includegraphics[width=\linewidth]{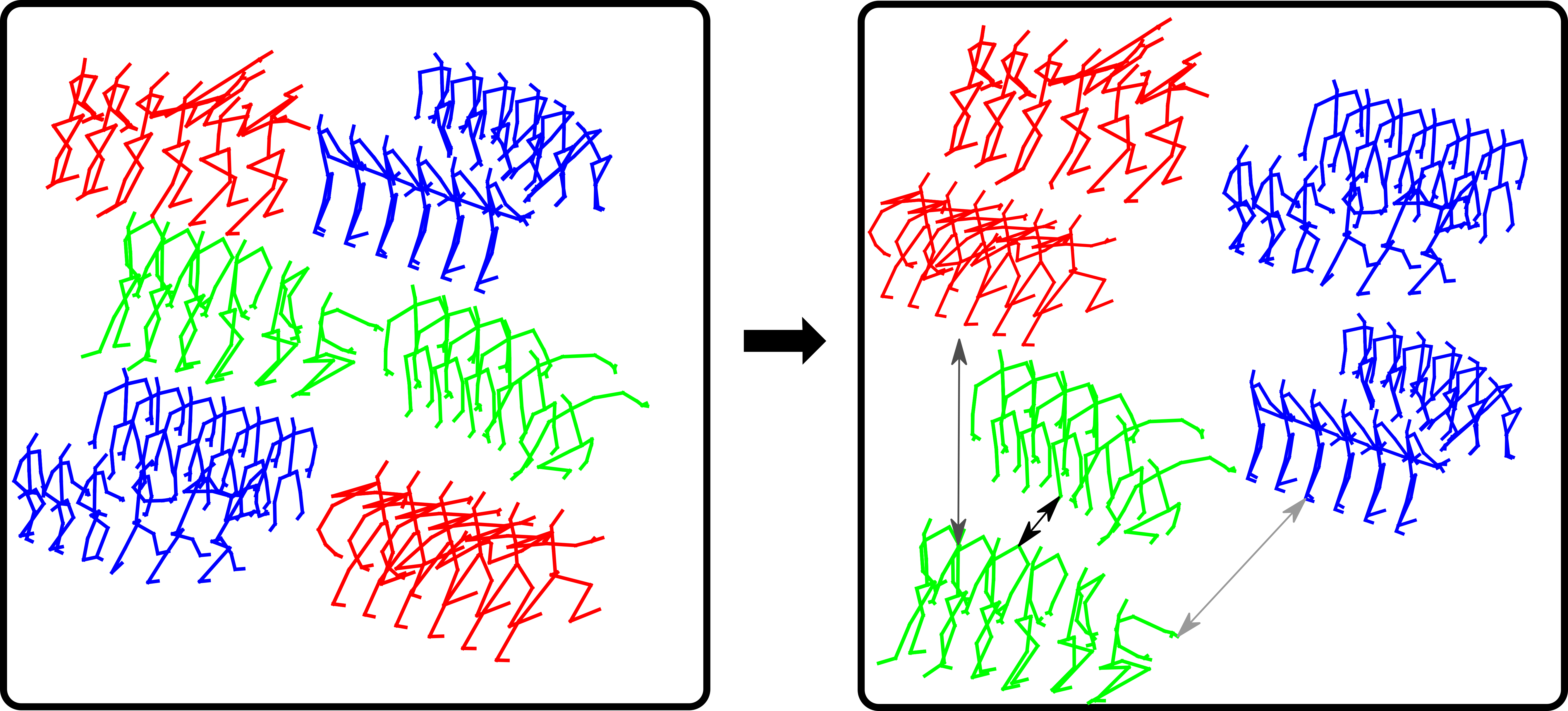}
    \caption{Illustrative example of our method. Prior to training a metric on the initial data, no class association could be formed given a skeleton sequence. After training our one-shot action recognition model, skeleton sequences can be encoded. A euclidean distance on the encoded sequence allows class association by finding the nearest neighbour in embedding space from a set of reference samples. The colors are encoding the following classes: \textcolor{red}{throw}, \textcolor{green}{falling}, \textcolor{blue}{grab other person's stuff}. Brighter arrow colors denote higher distance in embedding space.}
    \label{fig:motivation}
\end{figure}

Deep metric learning based approaches are popular for image ranking or clustering like face- or person re-identification \cite{schroff2015facenet, wojke2018deep}. They have proven to integrate well as an association metric, e.g.\, in person tracking settings to reduce the amount of id-switches \cite{wojke2018deep}.
Even though there are skeleton-based image representations for recognizing activities from skeleton sequences, they have only recently been used to learn a metric for one-shot action recognition \cite{memmesheimer2020signal}. \figname \ref{fig:motivation} shows an illustrative example of an application of our approach. 
The contributions of this paper are as follows:
\begin{itemize}
    \item We present a representation that reassembles skeleton sequences into images.
    \item We integrate the representation into a deep metric learning formulation to tackle the one-shot action recognition problem.
    \item We furthermore provide an evaluation of related skeleton-based image representations for one-shot action recognition.
    \item The source code to reproduce the results of this paper is made publicly available under \url{https://github.com/raphaelmemmesheimer/skeleton-dml}.
\end{itemize}

\section{Related Work}
\label{sec:related_work}

Action recognition is a broad research topic that varies not only in different modalities like image sequences, skeleton sequences, data by inertial measurement units but also by their evaluation protocols. Most common protocols are cross-view or cross-subject. More recently one-shot protocols have gained attention. As our approach focuses on skeleton-based one-shot action recognition we present related work from the current research state directly related to our method.
Skeleton based action recognition gained attention with the release of the Microsoft Kinect RGB-D camera. This RGB-D camera not only streamed depth and color images, but the SDK also supported the extraction of skeleton data. With the \textit{NTU RGB+D} dataset \cite{shahroudy2016ntu,liu2019ntu} a large scale RGB-D action recognition dataset that also contains skeleton sequences has been released. The progress made on this dataset gives a good indication of the performance of various skeleton-based action recognition approaches.


Because convolution neural architectures showed great performance in the image-classification domain, a variety of research concentrated on finding image-like representations for different research areas like speech recognition \cite{hershey2017cnn}.

Representations for encoding spatio-temporal information were explored in-depth for recognizing actions \cite{liu2017enhanced, caetano2019skelemotion}. They focus on a classification context by associating class labels with skeleton sequences in contrast to learning an embedding space. The idea of representing motion in image-like representations lead to serious alternatives to sequence classification approaches based on \textit{Recurrent Neural Networks} \cite{hu2018early} and \textit{Long Short Term Memory (LSTM)}  \cite{liu2017skeletonlstm}.
\citeauthordirect{wang2018action} presented joint trajectory maps. Viewpoints from each axis were set and encoded 3D trajectories for each of the three main axis views. A simple Convolutional Neural Network (CNN) architecture was used to train a classifier analyzing the joint trajectory maps. Occlusion could not be directly tackled, therefore the representation by \citeauthordirect{liu2017enhanced} added flexibility by fusing up to nine representation schemes in separate image channels. A similar representation has recently shown to be usable also for action recognition on different modalities and their fusion \cite{memmesheimer2020gimme}. \citeauthordirect{kim2017interpretable} on the other hand presented a compact and human-interpretable representation. Joint movement contributions over time can be interpreted.
Interesting to note is also the skeleton transformer by \citeauthordirect{li2017skeleton}. They employ a fully connected layer to transform skeleton sequences into a 2 dimensional matrix representation.

Yang \andothers \cite{yang2018action} present a joint order that puts joints closer together if their respective body parts are connected. 
It is generated by a depth-first tree traversal of the skeleton starting in the lower chest.
Skepxels are small $5\times5$-pixel segments containing the positions of all 25 skeleton joints in a random but fixed order. Liu \andothers \cite{liu2019skepxels} use this 2D structure as it is more easily captured by CNNs. Each sample of a sequence is turned into multiple sufficiently different Skepxels which are then stacked on top of each other. These Skepxels differ only in their joint permutation. The full Skepxel-image of a sequence of skeletons is assembled width-wise, without altering the joint permutation within one row of Skepxels.
Caetano \andothers \cite{caetano2019skelemotion} generate two images containing motion information in the form of an orientation and a magnitude. The orientation is defined by the angles between the motion vector and the coordinate axes. The angles are stored in the color channels of an image, with time in horizontal and the joints in TSSI order in vertical direction. The gray-scale magnitude image contains the euclidean norm of the motion vectors instead.



One-shot recognition in general aims at finding a method to classify new instances with a single reference sample. Possible approaches for solving problems of this category are metric learning \cite{wang2014learning,hoffer2015deep}, or meta-learning \cite{finn2017model}. In action recognition this means a novel action can be learned with a single reference demonstration of the action. In contrast to one-shot image classification, actions consist of sequential data. A single frame might not contain enough context to recognize a novel activity.

Along with the \textit{NTU RGB+D 120} dataset, \citeauthordirect{liu2019ntu} presented a one-shot action recognition protocol and corresponding baseline approaches. 
The \textit{Advanced Parts Semantic Relevance (APSR)} approach extracts features by using a spatio-temporal LSTM method. They propose a semantic relevance measurement similar to word embeddings. Body parts are associated with an embedding vector and a cosine similarity is used to calculate a semantic relevance score.
Sabater \andothers \cite{sabater2021one} presented a one-shot action recognition approach based on a Temporal Convolutional Network (TCN). After normalization of the skeleton stream, they calculate pose features and use the TCN for the generation of motion descriptors. The descriptors at the last frame, assumed to contain all relevant motion from the skeleton-sequence, are used to calculate the distances to the reference samples. Action classes are associated by thresholding the distances.
Our previous work on multi-modal one-shot action recognition \cite{memmesheimer2020signal} proposed to formulate the one-shot action recognition problem as a deep metric learning problem. Signals originating from various sensors are transformed into images and an encoder is trained using triplet-loss. The focus in that work was on showing the multi-modal applicability, whereas in this work we concentrate on skeleton-based one-shot action recognition.

\section{Approach}
\begin{figure}
    \vspace{0.06in}
    \centering
    \includegraphics[width=.6\linewidth]{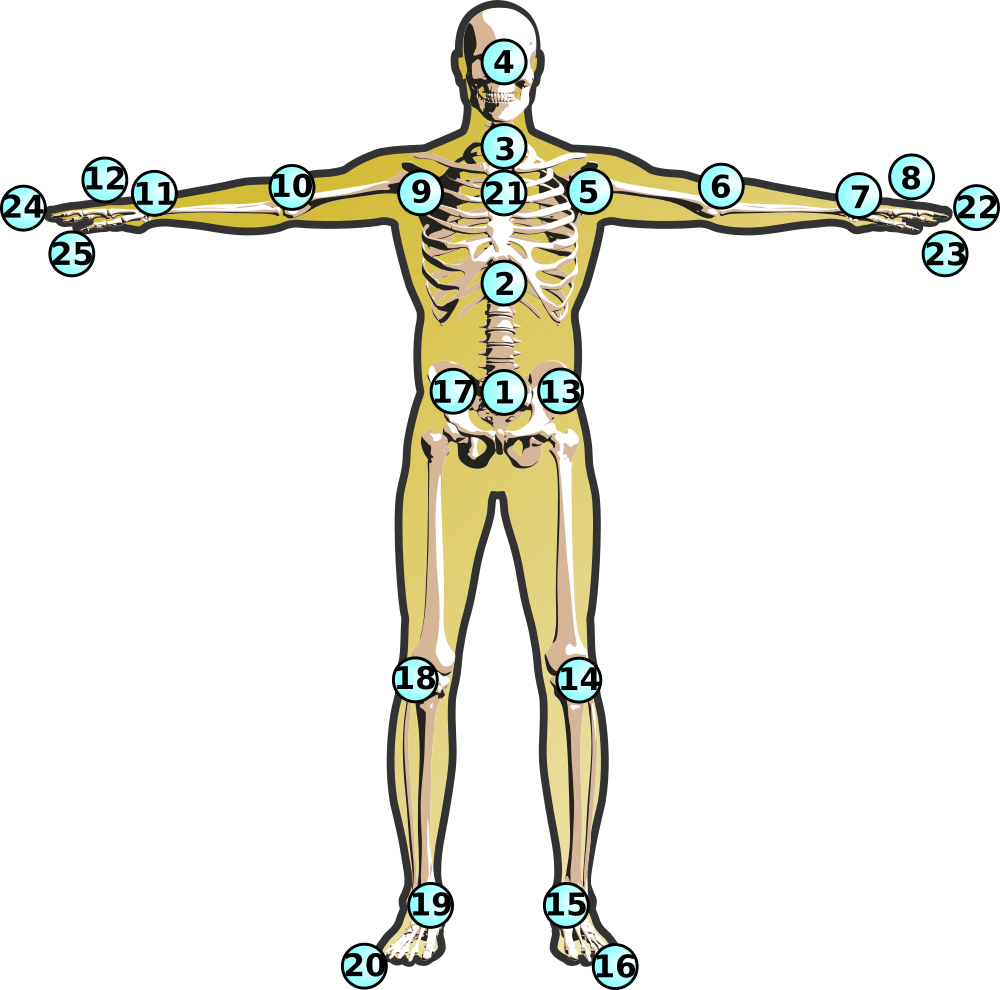}
    \caption{\textit{NTU RGB+D 120} skeleton joint positions.}
    \label{fig:skeleton}
\end{figure}


We propose a novel, compact image representation for skeleton sequences. Additionally we present an encoder model that learns to project said representations into a metric embedding space that encodes action similarity.

\subsection{Problem Formulation}
\label{sec:problem_formulation}

A standard approach for action recognition is trained on $\NumLabels$ classes, where the training and test sets share the same $\NumLabels$ classes. Thus a test set $\DatasetTest$ share the same classes as the training set $\Dataset$. In an one-shot action recognition setting $\NumLabels$ classes are known in a training set $\Dataset$, while the test set $\DatasetTest$ contains $\NumUnknownLabels$ novel classes, providing a single reference sample per class in an auxiliary set $\DatasetAuxiliary$, where $|\DatasetAuxiliary| = \NumUnknownLabels$. 
We consider the one-shot action recognition problem as a metric learning problem. 
Our goal is to train a feature embedding $\Feature = \Encode[\EncoderParams]{\Image}$ with parameters~$\EncoderParams$ which projects input images~$\Image\in\ImageSpace$, into a feature representation~$\Feature \in \FeatureSpace$. $H$ denotes the height of the image, $W$ denotes the width of the image in an RGB channel image and $d$ is the given target embedding vector size. The feature representation reflects minimal distances in embedding space for \textit{similar} classes. 
For defining the similarity we follow \cite{wang2019multi}, where the similarity of two samples $(\Image_i,\Feature_i)$ and $(\Image_j,\Feature_j)$ is defined as $\SimilarityElement_{ij} :=  <\Feature_i  , \Feature_j>$, where $<\cdot , \cdot>$ denotes the dot product, resulting in an $\NumData\times \NumData$ similarity matrix $\Similarities$.



\subsection{\approachname{} Representation}

\begin{figure}
    \vspace{0.06in}
    \centering
    \includegraphics[width=\linewidth]{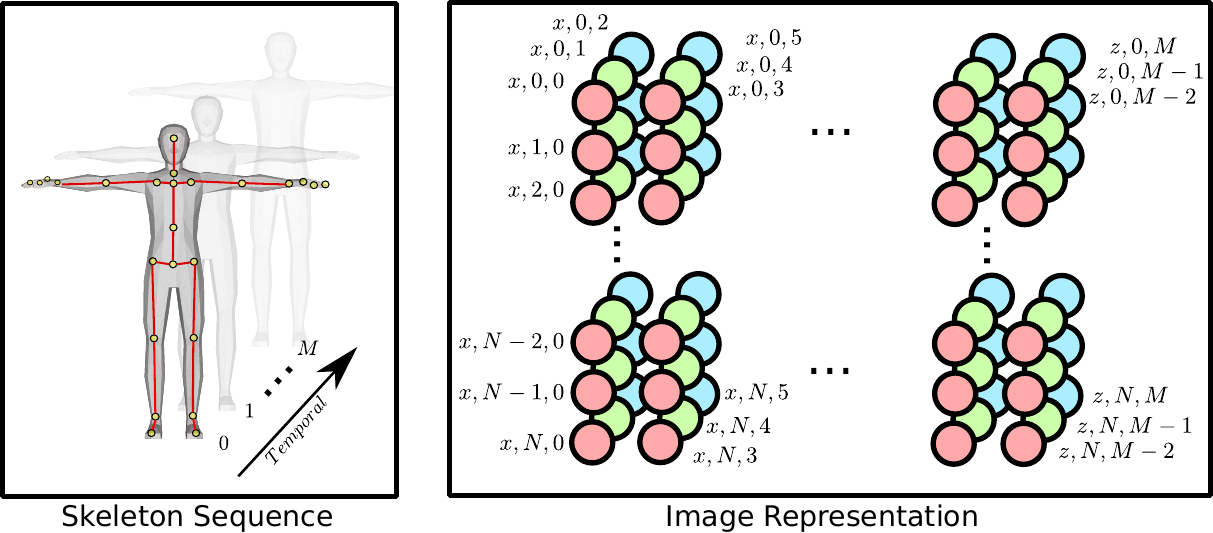}
    \caption{\approachname{} skeleton representation. $x$ and $z$  denote the skeleton joint component in joint space, the number of joints is reflected by $\NumJoints$, which relates to the height of the image $H$, the sequence length $\NumSequenceLength$ relates with the width of the image $W$. Note, instead of  projecting the temporal information throughout the width of the image, we project the joint space locally for each dimension and assemble the joint axis blocks over the width.}
    \label{fig:reindex_representation}
\end{figure}

\begin{figure*}
    \centering
    \includegraphics[width=.85\linewidth]{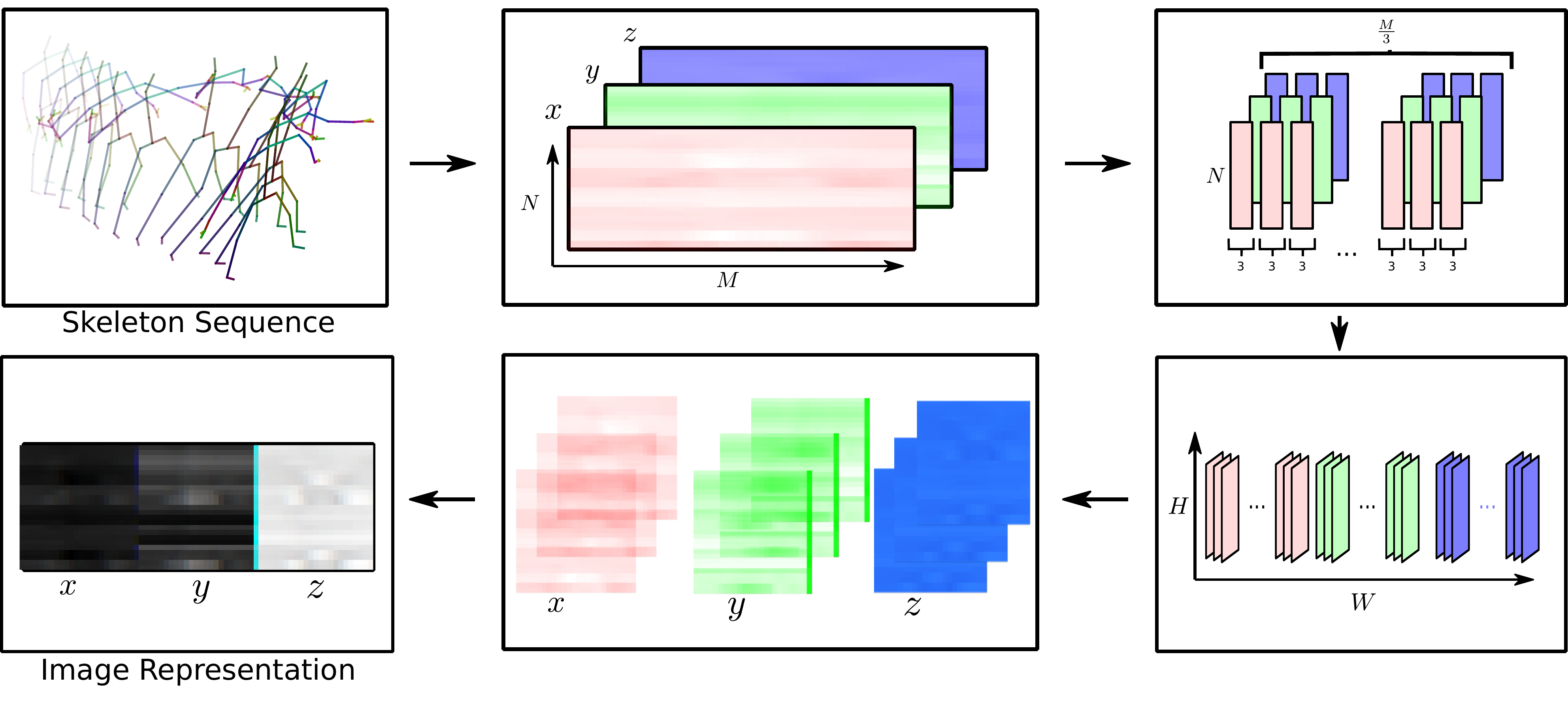}
    \caption{Exemplary representation for a throwing activity of the \textit{NTU-RGB+D 120} dataset. A skeleton-sequence serves an input and can be represented as an image directly \cite{du2015skeleton,memmesheimer2020gimme}. Our \approachname{} representation groups $x$-, $y$-, $z$ joint values locally in $\frac{\NumSequenceLength}{3}$ blocks per axis and assembles them into the final image representation. All axis blocks are laid out aside.}
    \label{fig:representation_example}
\end{figure*}

We encode skeleton sequences into an image representation. \figname \ref{fig:skeleton} shows the skeleton as contained in the NTU RGB+D 120 dataset. On a robotic system, these skeletons can be either directly extracted from the RGB-D camera \cite{zhang2012microsoft} or from a camera image stream using a human-pose estimation approach \cite{cao2019openpose}. 
The input in our case is a skeleton sequence matrix $\Signals \in \SignalSpace$ where each row vector represents a discrete joint sequence (for $\NumJoints$ joints) and each column vector represents a sample of all joint positions at one specific time step of a sequence length $\NumSequenceLength$. The matrix is transformed to an RGB image $\Image \in \ImageSpace$. Note, in contrast to \cite{memmesheimer2020signal, du2015skeleton} the joint space is not projected to the color channels but unfolded per axis separately like depicted in \figname \ref{fig:reindex_representation}, and \figname  \ref{fig:representation_example}.
This results in a dataset $\Dataset = \{(\Image_i, \Label_i)\}_{i=1}^{\NumData}$ of~$\NumData$ training images $\Image_{1, \dots, \NumData}$ with labels $\Label_i\in\{1, \dots, \NumLabels\}$. 
In contrast to the representations used for multimodal action recognition \cite{memmesheimer2020gimme} or skeleton based action recognition \cite{wang2018action,liu2017enhanced} the proposed representation is more compact. In comparison to \cite{du2015skeleton,memmesheimer2020signal} our representation separates the joint values for all axes as blocks over the width, keeping all joint values grouped locally together per axis. 
In \cite{memmesheimer2020signal} the color channels are used to unfold the joint values. As the skeleton-sequence is represented as an image, the model needs to be applied only to a single image for inference.

\subsection{Feature Extraction}

For better comparability between the approaches we use the same feature extraction method as previously proposed in \textit{SL-DML} \cite{memmesheimer2020signal}. Using a Resnet18 \cite{he2016deep} architecture allows us to train a model that converges fast and serves as a good feature extractor for the embedder. The low amount of parameters allows practical use for inference on autonomous mobile robots. Weights are initialized with a pre-trained model and are optimized throughout the training of the embedder. After the last feature layer we use a two-layer perceptron to transform the features to the given embedding size. The embedder is refined by the metric learning approach.

\subsection{Metric Learning}
\label{ssec:metric_learning}

\begin{figure*}
    \centering
    \includegraphics[width=.85\linewidth]{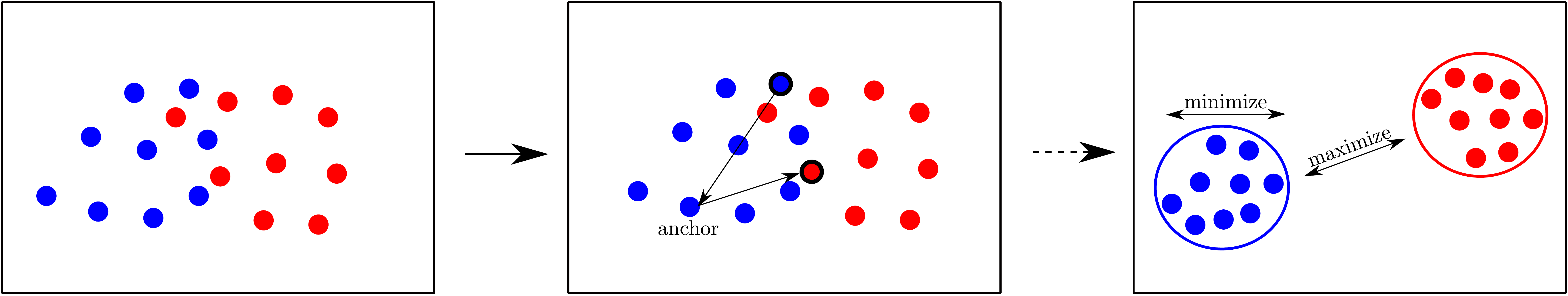}
    \caption{A possible intermediate state of the embeddings during the training process of two classes (left). During training, pairs, that are difficult to push apart in embedding space, are mined (middle). Given the blue anchor sample, the most difficult positive pair is the blue sample with the highest distance in embedding space. Similar, the closest red sample in embedding space is the corresponding negative sample. The overall goal is to separate the samples in embedding space (right) by minimizing the inter-class scatter and maximize the intra-class distance to the class centers in embedding space.}
    \label{fig:ms_sample}
\end{figure*}

Metric learning aims to learn a function to project an image into an embedding space, where the embedding vectors of similar samples are encouraged to be closer, while dissimilar ones are pushed apart from each other \cite{wang2019multi}. We use a \textit{Multi-Similarity-Loss} in combination with a \textit{Multi-Similarity-Miner} \cite{wang2019multi} for mining good pair candidates during training. Positive and negative pairs (by class label) that are assumed to be difficult to push apart in the embedding space are mined. \figname \ref{fig:ms_sample} gives a constructed example of how positive and negative pairs are mined.
Positive pairs are constructed by an anchor and positive image pair $\{\Image_\circ,\Image_\uparrow\}$ and its embedding $\Encode{\Image_\circ}$, preferring pairs with a low similarity in embedding space (high distance in embedding space) with the following condition:

\begin{equation}
\label{eq-select-pos}
\SimilarityElement^{+}_{\circ\uparrow} < \max_{k \neq \circ} \SimilarityElement_{\circ k} + \epsilon.
\end{equation}
Similar, if $\{\Image_\circ, \Image_\downarrow\}$ is a negative pair, the condition is:
\begin{equation}
\label{eq-select-neg}
\SimilarityElement^{-}_{\circ \downarrow} > \min_{k = \circ} \SimilarityElement_{\circ k} - \epsilon,
\end{equation}
where $k$ is a class label index and $\epsilon$ is a given margin.

Note, these conditions support the mining of hard pairs, i.e. a positive pair where the sample still has a high distance in embedding space and a negative pair that still has a low distance in embedding space. This forces sampling that concentrates on the hard pairs.
A set of positive images to an anchor image $\Image_\circ$ are denoted $\mathcal{P}_i$, analog, a set of negative images to $\Image_\circ$ are denoted $\mathcal{N}_i$. 

Given mined positive- and negative pairs allows us integration into the \textit{Multi-Similarity} loss, as derivated by \citeauthordirect{wang2019multi}:
\begin{equation}
    \begin{split}
    	\label{eq-MS}
    	\mathcal{L}_{MS} = \frac{1}{\NumData}\sum_{i=1}^\NumData  \bigg\{\frac{1}{\alpha}  { \log \big[1 + \sum_{k  \in \mathcal{P}_i } e^{-\alpha (\SimilarityElement_{ik} - \lambda)}}\big]  \\
    	+ \frac{1}{\beta }  { \log \big[1+ \sum_{k \in \mathcal{N}_i}
    		 e^{\beta (\SimilarityElement_{ik} - \lambda)} \big]} \bigg\},
    \end{split}   
\end{equation}
where $\alpha$, $\beta$ and $\lambda$ are fixed hyper-parameters.

\textcolor{black}{In contrast to \textit{SL-DML} we do not apply weighting to the classifier- and embedder loss, as no marginal improvement has been achieved in \cite{memmesheimer2020signal}.} After the model optimization, associating an action class to a query sample and set of reference samples is now reduced to a nearest-neighbor search in the embedding space. The classifier and encoder are jointly optimized.

\subsection{Implementation}
\label{ssec:implementation}

Our implementation is based on PyTorch \cite{Musgrave2019}, \cite{paszke2019pytorch}. 
We tried to avoid many of the metric learning flaws as pointed out by Musgrave \andothers \cite{musgrave2020metric} by using their training setup and hyperparameters where applicable.
Key differences are that we use a Resnet18 \cite{he2016deep} architecture and avoid the proposed four-fold cross validation for hyperparameter search in favour of better comparability to the proposed one-shot protocol on the \textit{NTU RGB+D 120} dataset \cite{liu2019ntu}. Note, we did not perform any optimization of the hyperparameters.
A batch size of 32 was used on a single Nvidia GeForce RTX 2080 TI with 11GB GDDR-6 memory. We trained for 100 epochs with initialized weights of a pre-trained Resnet18 \cite{he2016deep}. 
For the multi similarity miner we used an epsilon of 
$0.05$ 
and a margin of $0.1$ for the triplet margin loss. A 
RMSProp optimizer
with a learning rate of 
$10^{-6}$
was used in all optimizers. The embedding model outputs a 128 dimensional embedding. 

\section{Experiments}
\begin{figure}
    \centering
      \begin{tikzpicture}
            \begin{axis}[
                xlabel={\#Training Classes},
                ylabel={Accuracy},
                xmin=20, xmax=100,
                ymin=25, ymax=55,
                xtick={20,40,60,80,100},
                ytick={0,10,20,25,30,35,40,45,50,55},
                legend pos=south east,
                legend style={nodes={scale=0.8, transform shape}},
                ymajorgrids=true,
                grid style=dashed,
            ]
            \addplot[color=blue, mark=square] coordinates {(20,\results{apsr20})(40,\results{apsr40})(60,\results{apsr60})(80,\results{apsr80})(100,\results{apsr100})};\addlegendentry{APSR \cite{liu2019ntu}}
            
            \addplot[color=teal, mark=square] coordinates {(20,\results{sldml20})(40,\results{sldml40})(60,\results{sldml60})(80,\results{sldml80})(100,\results{sldml100})};\addlegendentry{SL-DML \cite{memmesheimer2020signal}}
            
            \addplot[color=red, mark=square] coordinates {(20,\results{sldmlreindex20})(40,\results{sldmlreindex40})(60,\results{sldmlreindex60})(80,\results{sldmlreindex80})(100,\results{sldmlreindex100})};\addlegendentry{Ours}
            \end{axis}
        \end{tikzpicture}
    \caption{Result graph for increasing auxiliary set sizes.}
    \label{fig:auxiliary_set_size_graph}
\end{figure}
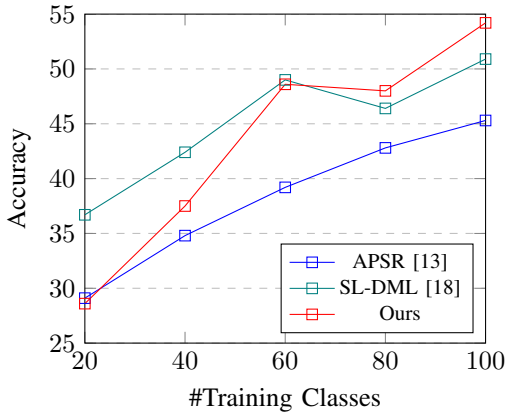

We used skeleton sequences from the \textit{NTU RGB+D 120} \cite{liu2019ntu} dataset for large scale one-shot action recognition.

The dataset is split into an auxiliary set, representing action classes that are used for training, and an evaluation set.  In the one-shot protocol the evaluation set does only contain novel actions. One sample of each test class serves as reference demonstration. This protocol is based on the one proposed by \cite{liu2019ntu} for the \textit{NTU RGB+D 120} dataset. 
First we trained a model on the auxiliary set. The resulting model transforms skeleton-sequences encoded as an image representation into embeddings for the reference actions and then for the evaluation actions. We then calculate the nearest neighbour from the evaluation embeddings to the reference embeddings. 
As the embeddings encode action similarity we can estimate to which reference sample the given test sample comes closest.
Beside the standard one-shot action protocol and experiments with dataset reduction, we give an ablation study that gives a hint on which combination of embedding size, loss, transformation and representation are yielding best results with our approach.
Further, we integrated various related skeleton-based image representations that have been previously proposed for action recognition into our one-shot action recognition approach to compare them. 

\subsection{Dataset}
\label{ssec:dataset}

\begin{figure*}[th] \centering$
  \vspace{0.06in}
  \begin{array}{cccc}
      \includegraphics[width=.2\linewidth]{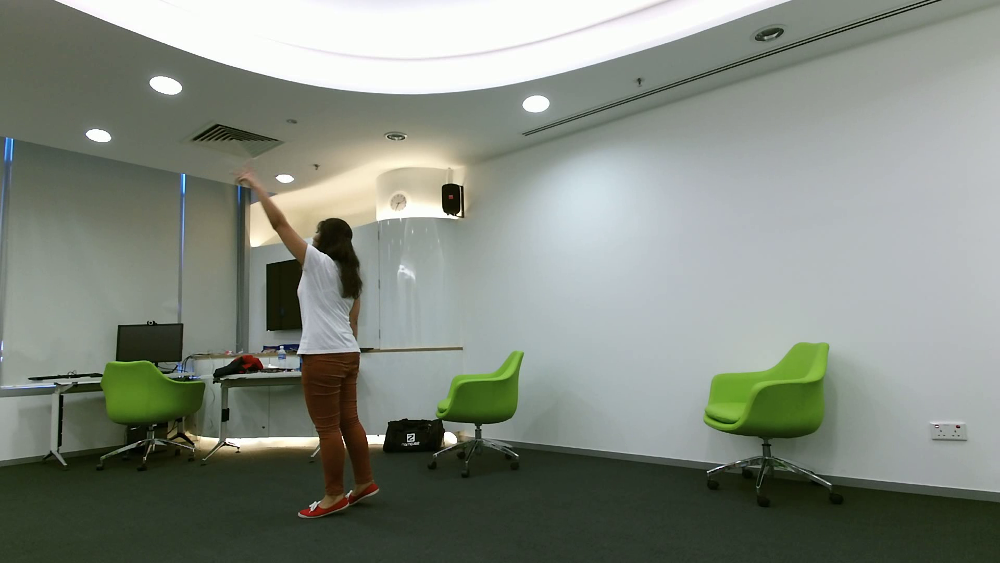} &
      \includegraphics[width=.2\linewidth]{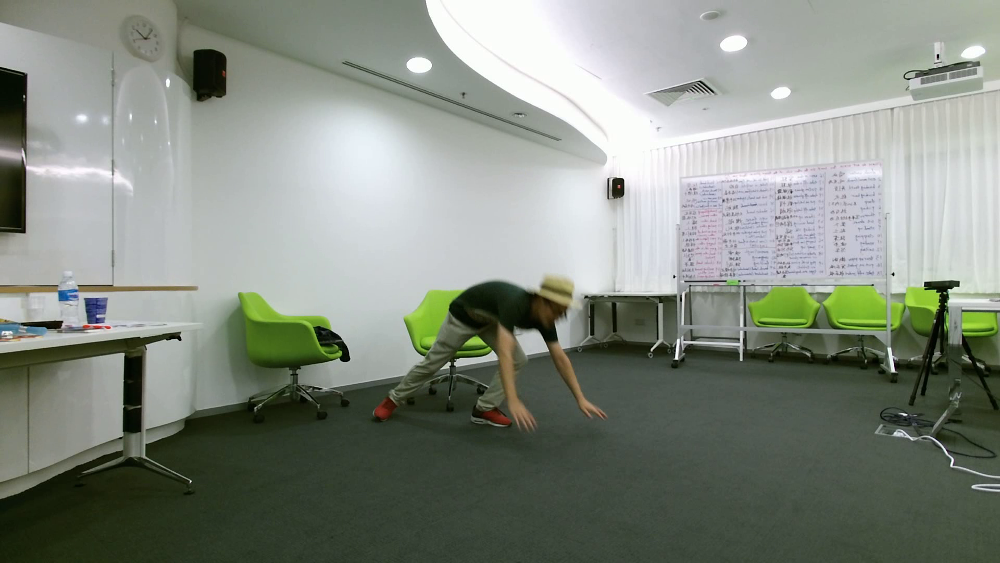} &
      \includegraphics[width=.2\linewidth]{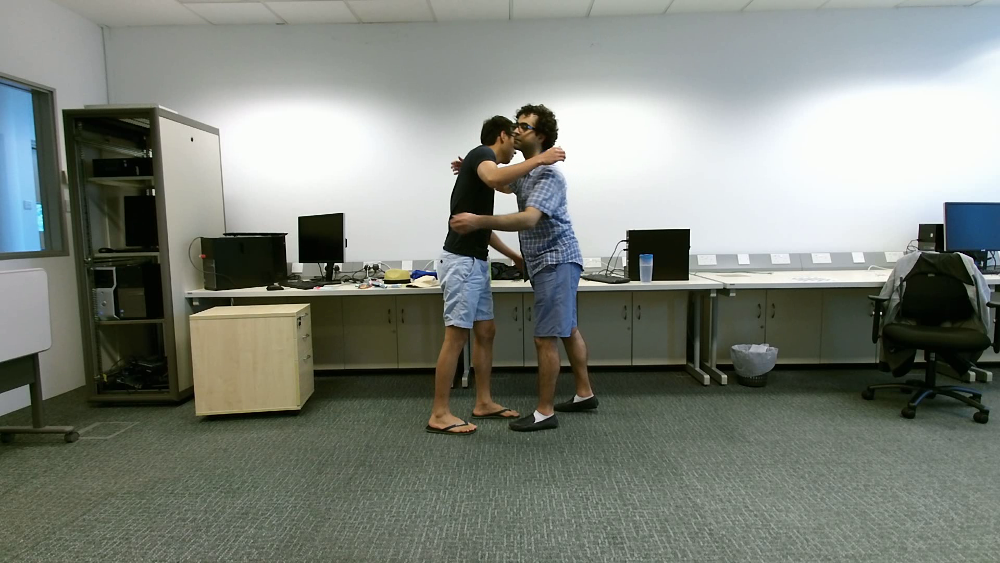} &
      \includegraphics[width=.2\linewidth]{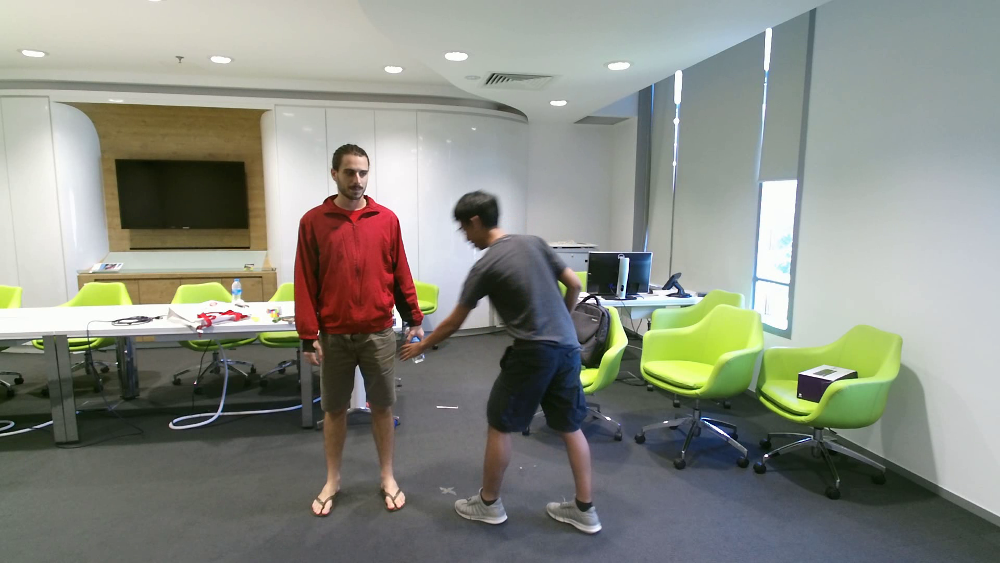} 
      \\
      \includegraphics[width=.2\linewidth]{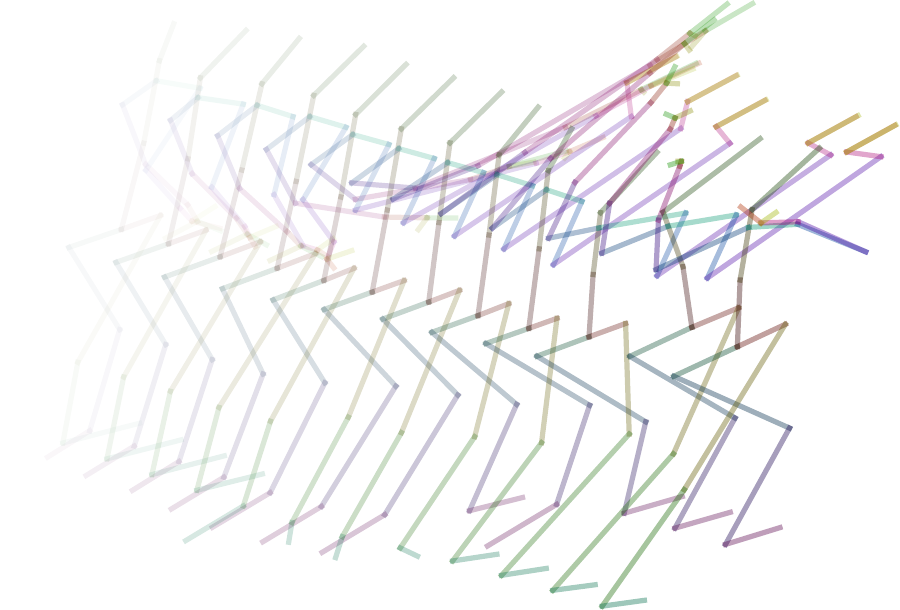} &
      \includegraphics[width=.2\linewidth]{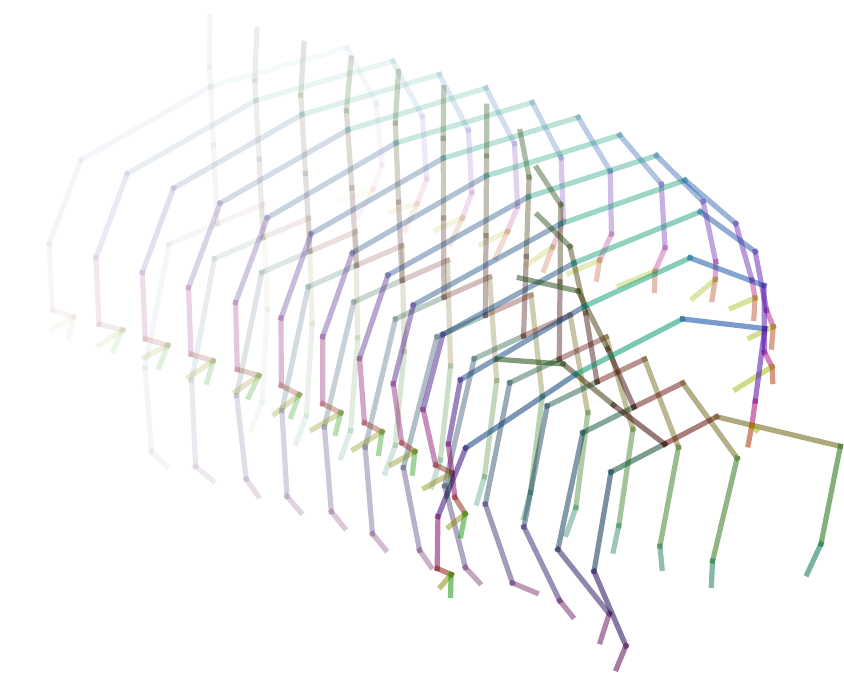} &
      \includegraphics[width=.2\linewidth]{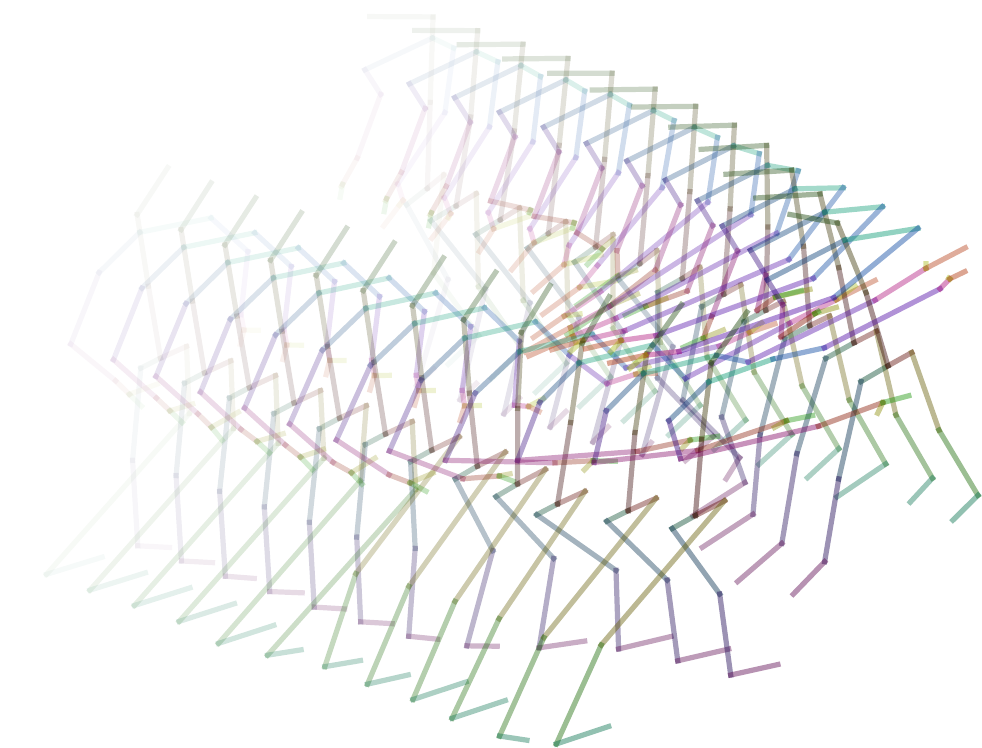} &
      \includegraphics[width=.2\linewidth]{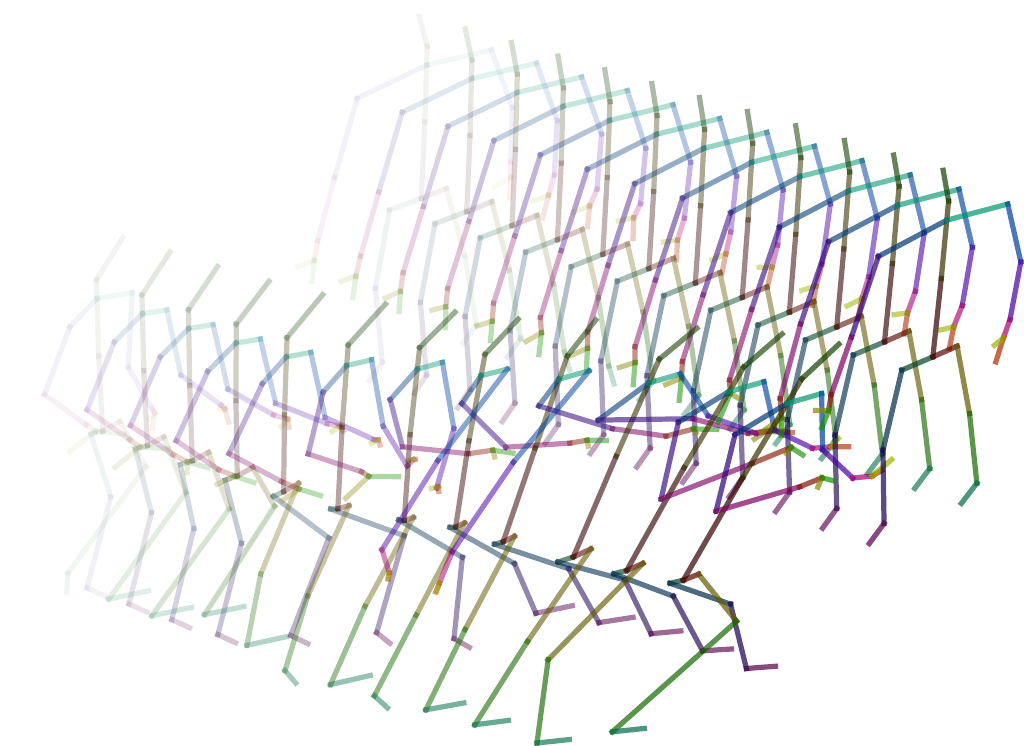} 
      \\
      \includegraphics[height=.08\linewidth]{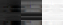} &
      \includegraphics[height=.08\linewidth]{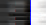} &
      \includegraphics[height=.12\linewidth]{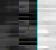}  &
      \includegraphics[height=.12\linewidth]{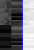} 
      \\
      \textrm{throw} & \textrm{falling}  & \textrm{hugging other person} & \textrm{grab other person's stuff}
  \end{array}$
\caption{From top to bottom: A RGB Frame, the corresponding skeleton sequences and the image representation of those sequences are shown. The latter is used in our one-shot action recognition approach. The first two sequences contain single person activities, whereas the remaining two contain two person interactions. The \textit{grab other person's stuff} sequence was shorter than the \textit{hugging other person} sequence.}
  \label{fig:examples}
\end{figure*}

\begin{table}[tb]
    \caption{One-shot action recognition results on the \textit{NTU RGB+D 120} dataset.}
	\begin{center}
        \small
		\begin{tabular}{lr}
			\toprule
            Approach                                       &  Accuracy [\%]\\
            \toprule
            Attention Network \cite{liu2017global}       &  41.0                   \\ 
            Fully Connected \cite{liu2017global}         &  42.1                   \\ 
			Average Pooling \cite{liu2017skeleton}       &  42.9                   \\ 
            APSR \cite{liu2019ntu}                       &  45.3          \\  
            TCN \cite{sabater2021one} & \results{sabater100} \\
            SL-DML \cite{memmesheimer2020signal}                  & \results{sldml100} \\ 
            Ours    & \textbf{\results{sldmlreindex100}} \\
            \bottomrule
		\end{tabular}
	\end{center}

	\label{tab:oneshot_results}
\end{table}

\begin{table}[tb]
	\caption{Results for different auxiliary training set sizes for one-shot recognition on the \textit{NTU RGB+D 120} dataset in \%.}
	\begin{center}
        \small
		\begin{tabular}{cccr}
			\toprule
			\#Train Classes   &  APSR \cite{liu2019ntu} & \textit{SL-DML} \cite{memmesheimer2020signal} &  Ours \\
            \toprule
             $20$               &     \results{apsr20}  & \textbf{\results{sldml20}}   &\results{sldmlreindex20}   \\ 
             $40$               &     \results{apsr40}  & \textbf{\results{sldml40}}   &\results{sldmlreindex40}  \\ 
             $60$               &     \results{apsr60}  & \textbf{\results{sldml60}}   &\results{sldmlreindex60}   \\ 
             $80$               &     \results{apsr80}  & \results{sldml80}   & \textbf{\results{sldmlreindex80}}   \\ 
			 $100$              &     \results{apsr100} & \results{sldml100}  & \textbf{\results{sldmlreindex100}}   \\ 
            \bottomrule
		\end{tabular}
	\end{center}

	\label{tab:ontshot2_results2}
\end{table}

The \textit{NTU RGB+D 120} \cite{liu2019ntu} dataset is a large scale action recognition dataset containing RGB-D image streams and skeleton estimates. 
The dataset consists of 114,480 sequences containing 120 action classes from 106 subjects in 155 different views.
We follow the one-shot protocol as described by the dataset authors. The dataset is split into two parts: an auxiliary set and an evaluation set. The action classes of the two parts are distinct. 100 classes are used for training, 20 classes are used for testing. The unseen classes and reference samples are documented in the accompanied dataset repository\footnote{\url{https://github.com/shahroudy/NTURGB-D}}. \textit{A1, A7, A13, A19, A25, A31, A37, A43, A49, A55, A61, A67, A73, A79, A85, A91, A97, A103, A109, A115} are previously unseen. As reference the demonstration for filenames starting with \textit{S001C003P008R001*} are used for actions with IDs below 60 and \textit{S018C003P008R001*} for actions with IDs above 60.
As no standard validation set is defined in the dataset paper we use the following classes during development for validation: \textit{A2, A8, A14, A20, A26, A32, A38, A44, A50, A56, A62, A68, A74, A80, A86, A92, A98, A104, A110, A116}.
One-shot action recognition results are given in \tabname \ref{tab:oneshot_results}. Like Liu \andothers \cite{liu2019ntu} we also experimented with the effect of the auxiliary set reduction. Results are given in \figname \ref{fig:auxiliary_set_size_graph} and \tabname \ref{tab:ontshot2_results2}. 
In addition we analyze different representations in \tabname\,\ref{tab:ablation_oneshot_ntu_representations} and the influence of different embedding vector sizes, metric losses and augmentations on two representations more detailed in \tabname\,\ref{tab:ablation_oneshot_ntu_embedding_size}.




\subsection{Training Set Size Reduction}
\label{ssec:training_set_reduction}

An interesting question that comes up when evaluating one-shot action recognition approaches is how much training classes are required to get a certain performance. 
Liu \andothers \cite{liu2019ntu} already proposed to evaluate the one-shot action recognition approach with varying training set sizes. Aligned with Liu \andothers \cite{liu2019ntu} we use training sets containing 20, 40, 60, 80 training classes while remaining a constant evaluation set size of 20. For practical systems, where only a limited amount of training data is available, this evaluation can give an important insight about which performance can be achieved with lower amounts of provided training data. It is also interesting to observe how an approach performs when adding more training data. \tabname \ref{tab:ontshot2_results2} and \figname \ref{fig:auxiliary_set_size_graph} give results for different training set sizes for \textit{SL-DML} \cite{memmesheimer2020signal}, \textit{APSR} \cite{liu2019ntu} and our \approachname{} approach, while remaining a static validation set. With just 20 training classes, our approach performs comparably to the \textit{APSR} approach. With a small amount of training classes the \textit{SL-DML} approach performs best. In our experiments \approachname{} performs better when providing a larger training set size. At a training set size of 60 classes, our approach performs comparably well to \textit{SL-DML}. With 80 classes in the training set our approach starts outperforming \textit{SL-DML}. It is interesting to note that, aligned with the results from \textit{SL-DML}, our approach seems to be confused by the 20 extra classes that are added to the 60 classes. 

\subsection{Ablation Study}
\label{ssec:ablation_study}

\begin{table}

    \caption{Ablation study for our proposed one-shot action recognition with different representations, embedding sizes, losses and augmentations. Results are given for a training over 200 epochs. Units are in \%.}
	\begin{center}
        \small
		\begin{tabular}{lrrrrr}
		    \toprule
            Representation                      & 128 & 256 & 512 & Transform & Loss \\
            \toprule
            SL-DML \cite{memmesheimer2020signal} & \textcolor{black}{\results{sldmlswap_transform_default_emb_128_loss_ms_miner_ms_alpha_1.0_beta_0.0}} & \results{sldmlswap_transform_default_emb_256_loss_ms_miner_ms_alpha_1.0_beta_0.0} &
            \results{sldmlswap_transform_default_emb_512_loss_ms_miner_ms_alpha_1.0_beta_0.0} & None & MS \\
            SL-DML \cite{memmesheimer2020signal} &
            \results{sldmlswap_transform_default_emb_128_loss_ms_miner_triplet_margin_alpha_1.0_beta_0.0} & \results{sldmlswap_transform_default_emb_256_loss_ms_miner_triplet_margin_alpha_1.0_beta_0.0} & \results{sldmlswap_transform_default_emb_512_loss_ms_miner_triplet_margin_alpha_1.0_beta_0.0} & None & TM  \\
            SL-DML \cite{memmesheimer2020signal} & \results{sldmlswap_transform_rotation_5_emb_128_loss_ms_miner_ms_alpha_1.0_beta_0.0} & \results{sldmlswap_transform_rotation_5_emb_256_loss_ms_miner_ms_alpha_1.0_beta_0.0} &
            \results{sldmlswap_transform_rotation_5_emb_512_loss_ms_miner_ms_alpha_1.0_beta_0.0} &  Rot & MS  \\
            SL-DML \cite{memmesheimer2020signal} &
            \results{sldmlswap_transform_rotation_5_emb_128_loss_ms_miner_triplet_margin_alpha_1.0_beta_0.0} & \results{sldmlswap_transform_rotation_5_emb_256_loss_ms_miner_triplet_margin_alpha_1.0_beta_0.0} & \results{sldmlswap_transform_rotation_5_emb_512_loss_ms_miner_triplet_margin_alpha_1.0_beta_0.0} & Rot & TM   \\
            Ours &
            \results{sldmlreindex_transform_default_emb_128_loss_ms_miner_ms_alpha_1.0_beta_0.0} & \results{sldmlreindex_transform_default_emb_256_loss_ms_miner_ms_alpha_1.0_beta_0.0} &
            \results{sldmlreindex_transform_default_emb_512_loss_ms_miner_ms_alpha_1.0_beta_0.0} & None & MS  \\
            Ours &
            \results{sldmlreindex_transform_default_emb_128_loss_ms_miner_triplet_margin_alpha_1.0_beta_0.0} & \results{sldmlreindex_transform_default_emb_256_loss_ms_miner_triplet_margin_alpha_1.0_beta_0.0} & \results{sldmlreindex_transform_default_emb_512_loss_ms_miner_triplet_margin_alpha_1.0_beta_0.0} & None & TM\\
            Ours &
            \results{sldmlreindex_transform_rotation_5_emb_128_loss_ms_miner_ms_alpha_1.0_beta_0.0} & \textcolor{black}{\results{sldmlreindex_transform_rotation_5_emb_256_loss_ms_miner_ms_alpha_1.0_beta_0.0}}&
            \textbf{\results{sldmlreindex_transform_rotation_5_emb_512_loss_ms_miner_ms_alpha_1.0_beta_0.0}} & Rot & MS \\
            Ours &
            \textcolor{black}{\results{sldmlreindex_transform_rotation_5_emb_128_loss_ms_miner_triplet_margin_alpha_1.0_beta_0.0}} & 
            \results{sldmlreindex_transform_rotation_5_emb_256_loss_ms_miner_triplet_margin_alpha_1.0_beta_0.0} & \textcolor{black}{\results{sldmlreindex_transform_rotation_5_emb_512_loss_ms_miner_triplet_margin_alpha_1.0_beta_0.0}} & Rot& TM  \\
            
            \bottomrule
		\end{tabular}
	\end{center}

	\label{tab:ablation_oneshot_ntu_embedding_size}
\end{table}

To distill the effects of the components we report their individual contributions. We examine influence of the representation, augmentation method and different resulting embedding vector sizes. 
Inspired by Roth \andothers \cite{roth2020revisiting} we experiment with different embedding vector sizes of 128, 256, 512. In addition we included the \textit{SL-DML} representation, compare a Triplet Margin loss (TM) and a Multi-Similarity loss (MS) and included an augmentation with random rotations of 5$^{\circ}$. In total 24 models were trained for this ablation study. We trained these models for 200 epochs as we expected longer convergence due to the additional augmented data. Results are given in \tabname \ref{tab:ablation_oneshot_ntu_embedding_size}. In the table we highlight important results. We highlight interesting results by different colors in the table (best result \textcolor{black}{without augmentation (\results{sldmlswap_transform_default_emb_128_loss_ms_miner_ms_alpha_1.0_beta_0.0}\%)}, \textcolor{black}{embedding size of 128 (\results{sldmlreindex_transform_rotation_5_emb_128_loss_ms_miner_triplet_margin_alpha_1.0_beta_0.0}\%)}, \textcolor{black}{embedding size of 256 (\results{sldmlreindex_transform_rotation_5_emb_256_loss_ms_miner_ms_alpha_1.0_beta_0.0}\%)}, \textcolor{black}{TM loss (\results{sldmlreindex_transform_rotation_5_emb_512_loss_ms_miner_triplet_margin_alpha_1.0_beta_0.0}\%)}, overall, MS loss, augmentation, embedding size of 512 (\results{sldmlreindex_transform_rotation_5_emb_512_loss_ms_miner_ms_alpha_1.0_beta_0.0}\%)). 
For \textit{SL-DML} the augmentation had a positive influence with higher embedding vector sizes of 512. Whereas the augmentation with embedding sizes of 128 only improved with the TM loss. With the MS loss and a low embedding size the augmentation did lower the result. For our \approachname{} representation the augmentation improved the results throughout the experiments for both losses. The best results without augmentation were achieved by the \textit{SL-DML} representation with an embedding vector of size 128 and a MS loss.
The overall best results were achieved with a MS loss and embedding vector size of 512 and augmentation by rotation using the \approachname{} representation, which improved the results of \skeletondmlaugmentedimpro{} over our approach under a comparable training setup as \textit{SL-DML}.

\subsection{Comparison with Related Representations}

\begin{table}
    \vspace{0.06in}
    \caption{Ablation study for different representations.}
	\begin{center}
        \small
		\begin{tabular}{lr}
		   \toprule
            Representation                      & Accuracy [\%] \\
            \toprule
            Skepxel \cite{liu2019skepxels} & \results{skepxel100}\\
            SkeleMotion Orientation \cite{caetano2019skelemotion} & \results{caetanoorientation100}\\
            SkeleMotion MagnitudeOrientation \cite{caetano2019skelemotion} & \results{caetanomagnitudeorientation100}\\
            TSSI \cite{yang2018action} & \results{tssi100}\\
            Gimme Signals \cite{memmesheimer2020gimme} & \results{gimme_signals100}\\
            SkeleMotion Magnitude \cite{caetano2019skelemotion} & \results{caetanomagnitude100}\\
            SL-DML \cite{memmesheimer2020signal} & \results{sldml100}\\
            Ours & \textbf{\results{sldmlreindex100}}\\
             
            \bottomrule
		\end{tabular}
	\end{center}

	\label{tab:ablation_oneshot_ntu_representations}
\end{table}

To support the effectiveness of our proposed representation in a metric learning setting we compare against other skeleton-based image representations. We use the publicly avail able implementation for the \textit{SkeleMotion} \cite{caetano2019skelemotion}, \textit{SL-DML} \cite{memmesheimer2020signal}, Gimme Signals \cite{memmesheimer2020gimme} and re-implementations of the \textit{TSSI} \cite{yang2018action} and \textit{Skepxels} \cite{liu2019skepxels} representations to integrate them into our metric learning approach. These representations have been described in 
\secname \ref{sec:related_work} more detailed. 

 The overall training procedure was identical as all models were trained with the parameters described in \secname  \ref{ssec:implementation}. The experiment only differed in the underlying representation. Results for the representation comparison  are given in \tabname \ref{tab:ablation_oneshot_ntu_representations}.
While most of the representations initially target action recognition and are not optimized for one-shot action recognition, they are still good candidates for integration in our metric learning approach. 
We did not re-implement the individual architecture proposed by the different representations but decided to use the Resnet18 architecture for better comparability.

Our \approachname{} approach shows best performance followed by \textit{SL-DML}. The \textit{SkeleMotion} Magnitude \cite{caetano2019skelemotion} representation transfers well from an action recognition setting to a one-shot action recognition setting. Interesting to note is that the \textit{SkeleMotion} Orientation  \cite{caetano2019skelemotion} representation, while achieving comparable results in the standard action recognition protocol, performs 10\% worse than the same representation encoding the magnitude of the skeleton joints. 
\textcolor{black}{An early fusion of Magnitude and Orientation on a representation level did not improve the Skelemotion representation but yields a result in between both representations. Similar observations have been made in \cite{memmesheimer2020signal} by the fusion of inertial and skeleton sequences. The lower performing modality adds uncertainty to the resulting model in our one-shot setting.}

A \textit{UMAP} embedding of all evaluation samples is shown in \figname \ref{fig:umap_embedding} for our \approachname{} approach. 
Our approach shows better capabilities in distinguishing the actions \textcolor{throw}{throw} and \textcolor{arm_circles}{arm circles}. In our approach those clusters can be separated quite well whereas \textit{SL-DML} struggles to discriminate those two classes.

\subsection{Result Discussion}
\label{ssec:results}

\begin{figure}
    \centering
    $  \begin{array}{cc}
      \includegraphics[width=.7\linewidth]{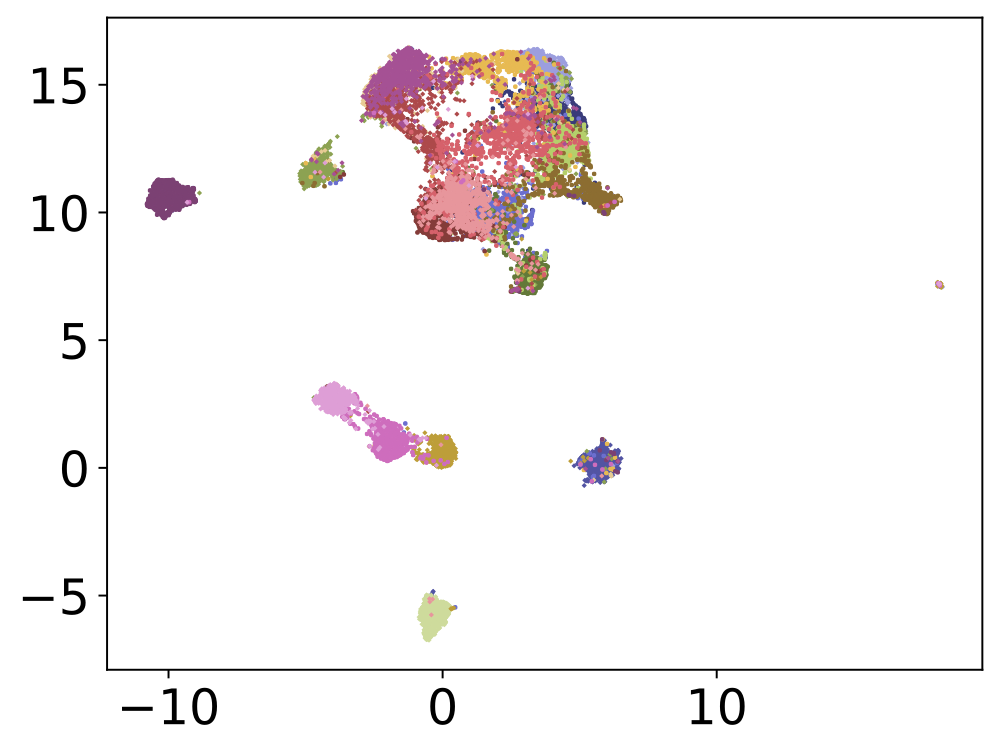} &
     \end{array}$
    \caption{\textit{UMAP} embedding visualization four our approach. 
    Classes are: drink water \textcolor{drink_water}{$\bullet$},
throw \textcolor{throw}{$\bullet$},
tear up paper \textcolor{tear_up_paper}{$\bullet$},
take off glasses \textcolor{take_off_glasses}{$\bullet$},
reach into pocket \textcolor{reach_into_pocket}{$\bullet$},
pointing to something with finger \textcolor{pointing_to_something_with_finger}{$\bullet$},
wipe face \textcolor{wipe_face}{$\bullet$},
falling \textcolor{falling}{$\bullet$},
feeling warm \textcolor{feeling_warm}{$\bullet$},
hugging other person \textcolor{hugging_other_person}{$\bullet$},
put on headphone \textcolor{put_on_headphone}{$\bullet$},
hush (quite) \textcolor{hush_(quite)}{$\bullet$},
staple book \textcolor{staple_book}{$\bullet$},
sniff (smell) \textcolor{sniff_(smell)}{$\bullet$},
apply cream on face \textcolor{apply_cream_on_face}{$\bullet$},
open a box \textcolor{open_a_box}{$\bullet$},
arm circles \textcolor{arm_circles}{$\bullet$},
yawn \textcolor{yawn}{$\bullet$},
grab other person’s stuff \textcolor{grab_other_persons_stuff}{$\bullet$},
take a photo of other person \textcolor{take_a_photo_of_other_person}{$\bullet$}.}
    \label{fig:umap_embedding}
\end{figure}

We evaluated our approach in an extensive experiment setup. Aside from lower performance on lower amounts of classes for training our approach outperformed other approaches. For fair comparison we report the result of \skeletondmlimpro{} over \textit{SL-DML} for training with 100 epochs and without augmentation, as under these conditions the \textit{SL-DML} result was reported. With augmentation and training for 200 epochs we could improve the baseline for \skeletondmlaugmentedimprosldml. Our approach learns to learn an embedding model that captures semantic relevance from joint movements well. E.g. \approachname{} differentiates well between activities that primarily contain hand- or leg-movements. Interactions between multiple person and single person activities are also separated well. Activities to which similar joint movements contribute to are still challenging. These are the activities that are formed by the main cluster in \figname \ref{fig:umap_embedding}.

\section{Conclusion}
We presented a one-shot action recognition approach based on the transformation of skeleton sequences into an image representation. On the image representations an embedder is learned which projects the images into an embedding vector. Distances between encoded actions reflect semantic similarities. Actions can then be classified, given a single reference sample, by finding the nearest neighbour in embedding space.
In an extensive experiment setup we compared different representations, losses, embedding vector sizes and augmentations.
Our representation remains flexible and yields improved results over \textit{SL-DML}. Additional augmentation by random 5 degree rotations have shown to further improve the results.
We found the overall approach of transforming skeleton sequences into image representations for one-shot action recognition by metric learning a promising idea that allows future research into various directions like finding additional 
representations, augmentation methods or mining and loss approaches. Especially in robot applications one-shot action recognition approaches have the potential to improve human robot interaction by allowing robots to adapt to unknown situations. The required computational cost for our approach is low, as only a single image representations of the skeleton-sequence needs be embedded by a comparably slim Resnet18-based embedder.

\printbibliography



\end{document}